\documentclass[letterpaper, 10 pt, conference]{ieeeconf}  

\IEEEoverridecommandlockouts                              

\usepackage{mathtools}
\usepackage{amsmath}
\usepackage{pdfpages}
\usepackage{cite}
\usepackage{hyperref}
\usepackage[normalem]{ulem}
\usepackage{amssymb}
\usepackage{adjustbox}
\usepackage{booktabs}
\usepackage{tabularx}

\usepackage{subcaption}

\usepackage[ruled,vlined]{algorithm2e}

\usepackage{graphicx}
\usepackage{etoolbox}
\newcommand{\insertfig}{
\includegraphics[width=2.01\columnwidth]{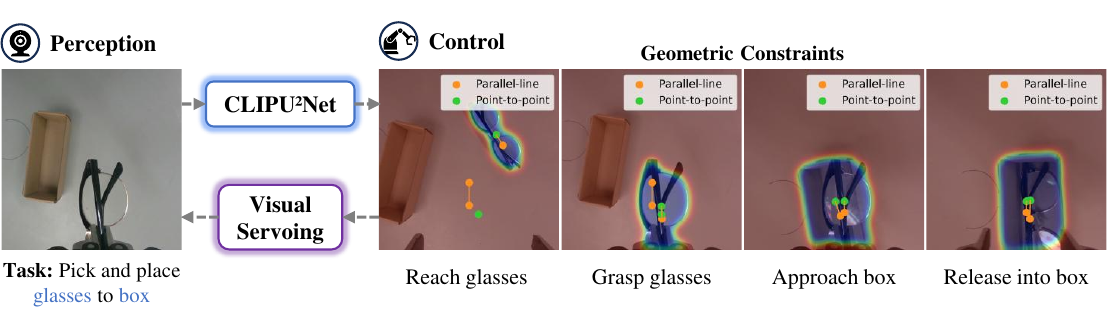}%
\captionof{figure}{To solve robot manipulation tasks in real-world environments, CLIPU$^2$Net is first employed to segment regions most relevant to the target specified by referring language. Geometric constraints are then applied to the segmented region, generating context-relevant motions for uncalibrated image-based visual servoing (UIBVS) control.}
}

\makeatletter
\apptocmd{\@maketitle}{\centering\insertfig}{}{}
\makeatother

\title{\LARGE \bf
Robot Manipulation in Salient Vision through Referring Image Segmentation and Geometric Constraints

}

\author{Chen Jiang$^{\dagger}$, Allie Luo$^{\dagger}$ and Martin Jagersand$^{\dagger}$
\thanks{$^{\dagger}$Authors are with Department of Computing Science,
        University of Alberta, Edmonton AB., Canada, T6G 2E8.
        { 
           \tt\small \{cjiang2, luo3, mj7\}@ualberta.ca
        }
        }%
}

\begin{document}

\maketitle
\thispagestyle{empty}
\pagestyle{empty}


\begin{abstract}
In this paper, we perform robot manipulation activities in real-world environments with language contexts by integrating a compact referring image segmentation model into the robot's perception module. First, we propose CLIPU$^2$Net, a lightweight referring image segmentation model designed for fine-grain boundary and structure segmentation from language expressions. Then, we deploy the model in an eye-in-hand visual servoing system to enact robot control in the real world. The key to our system is the representation of salient visual information as geometric constraints, linking the robot's visual perception to actionable commands. Experimental results on 46 real-world robot manipulation tasks demonstrate that our method outperforms traditional visual servoing methods relying on labor-intensive feature annotations, excels in fine-grain referring image segmentation with a compact decoder size of 6.6 MB, and supports robot control across diverse contexts.

\end{abstract}

\section{Introduction}
An eye-in-hand visual servoing system \cite{jagersand1997experimental, gridseth2016vita} depends on the flow of information from perception to control. In the perception phase, the system captures visual data with the eye-in-hand camera and processes this data to understand the manipulation context. The contextual data is then used in the control phase, where motor commands are generated to perform precise actions. For a robot to execute meaningful commands, it is necessary to interpret the evolving contexts of its workspace. However, the inherent movements of the eye-in-hand camera can cause visibility issues, where portions of objects may be obscured or fall outside the field of view, introducing significant challenges in perception.

Previous studies \cite{abolghasemi2019pay, jiang2020understanding, zha2021contrastively, an2024skill, yoshida2024text} have investigated the relationship between salient visual features and natural language in hand-eye cooperation, identifying regions that correlate with affordances. Recent advancements in large language models (LLMs) and vision-language models (VLMs) offer even more promising methods in generating point-based affordance representations from pixels for control \cite{liu2024moka}. However, the significant computational costs associated with LLMs pose difficulties for real-time robot control. Additionally, the use of points may overlook important details, such as fine boundaries and structures of the targets, making them less suitable for tasks requiring precise alignments. Consequently, exploring more compact models that balance between good affordance representations and computational efficiency becomes increasingly attractive. Can more compact models capture salient visual features from referring language, while enabling efficient robot manipulation?

To this end, extended from our previous work \cite{jiang2024clipunetr}, we further explore how salient visual information interacts with robot control in real-world environments. We summarize our contributions as follows:

\begin{itemize}
\item We introduce CLIPU$^2$Net, a new CLIP-driven model that delivers fine-grained referring image segmentation, with a compact decoder size of just 6.6MB.

\item We approach robot manipulation in salient vision by framing it as a visual task specification problem, using CLIPU$^2$Net and geometric constraints of points and lines to translate tasks into motions.

\item Through experiments on 46 real-world tasks with varying appearances and contexts, we demonstrate the effectiveness of CLIPU$^2$Net-inferred geometric constraints as universal visual representations for motions.

\end{itemize}

\section{Related Work}
\subsection{Salient Visual Features in Robot Control}
Salient visual features have been extensively studied in robotics embodying cues. Particularly, classical methods like Vita \cite{jagersand1997experimental, gridseth2016vita} relied on humans to annotate the cues as geometric constraints from parts of the objects. Jin et al \cite{jin2022generalizable} proposed to compose geometric constraints from dense visual descriptors into graph-based task functions. On the other hand, salient visual features can be learned unsupervised manifested as affordance signals, suggesting direct correlations between robotic actions and contexts. Vision-based approaches \cite{yang2019learning, jiang2020understanding, zha2021contrastively, bahl2023affordances, yoshida2024text} involved using vision-language or contrastive models to infer cues from demonstration videos, while control-based approaches \cite{zeng2021transporter, shridhar2022cliport, collins2024forcesight} learned to predict affordance scoring functions and actions. In particular, Bahl and Yoshida et al \cite{bahl2023affordances, yoshida2024text} proposed to learn affordances from human videos, and deployed for robot control. Collin et al \cite{collins2024forcesight} learned to predict affordance heatmaps from joint vision, language and force inputs and enacted control. Still, how to effectively bridge from salient features to robot control remains challenging.

\subsection{Vision-language Models in Robot Control}
The uses of Vision-Language Models (VLMs) and Large Language Models (LLMs) have gained popularity due to their strong contextual reasoning capabilities. Methods such as Grounded Decoding \cite{huang2024grounded}, MOKA \cite{liu2024moka}, AffordanceLLM \cite{qian2024affordancellm}, ManipLLM \cite{li2024manipllm}, and OVAL-Prompt \cite{tong2024ovalprompt} have demonstrated that LLMs can extract robust point-based affordance representations and be utilized as embodied agents. However, a significant drawback of LLM-based control systems is their high computational cost. For example, MOKA employed Grounded SAM \cite{ren2024grounded} and Octo \cite{mees2024octo} to localize pixels and generate motion cues, while Grounded Decoding relied on CLIPort \cite{shridhar2022cliport} for language-conditioned actions. These compositional strategies further amplify the computational demands of the control system.

In contrast, more compact approaches have been explored using smaller-scale models. Studies like Shridhar and Zhuo et al \cite{shridhar2022cliport, ijcai2024p201} employed CLIP \cite{radford2021learning} to localize center attention or affordance cues for policy learning. However, these learning-based methods often require a large number of demonstrations per task for training. Jiang et al \cite{jiang2024clipunetr} utilized CLIPUNetr, a referring image segmentation model, to segment targets before applying UIBVS for control, though their work was limited to the reach-and-grasp context only.

\section{Methodology}
\begin{figure}[h!]
	\centering 
	\includegraphics[width=0.75\columnwidth]{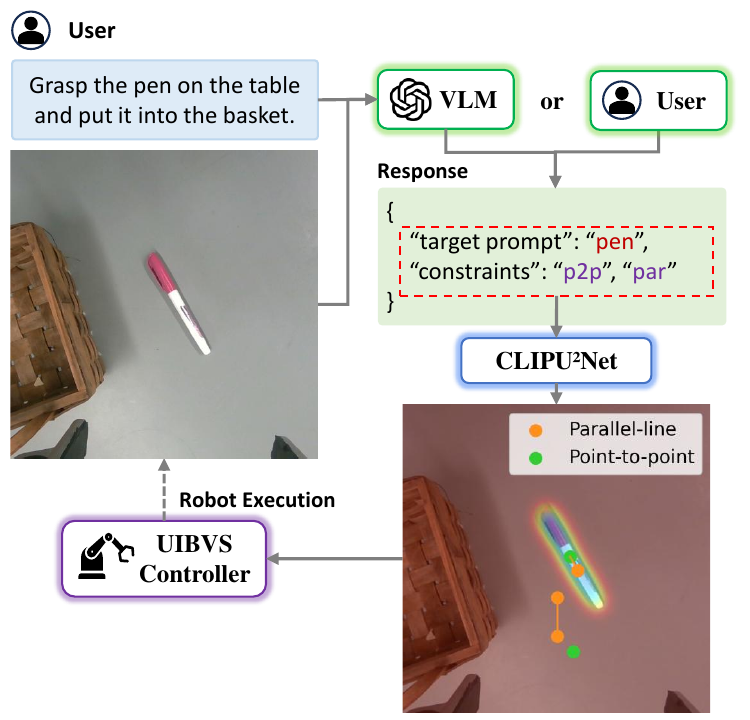}	
	\caption{Overview of the system to enact real-world robot control with CLIPU$^2$Net and UIBVS control.} 
	\label{fig:pipeline}%
\end{figure}

We consider single-arm manipulation with an eye-in-hand camera. The goal is to segment regions related to the context specified by a referring expression in natural language, and compute geometric constraints using geometric points and lines for UIBVS control. Extended from \cite{jiang2024clipunetr}, the overview of the system is presented in Figure \ref{fig:pipeline}. To enable the system, we first introduce CLIPU$^2$Net, a compact model used to perform referring image segmentation. Then, we describe the process to compose geometric constraints in salient vision. 

\begin{figure*}[ht!]
\centering
\includegraphics[width=1.75\columnwidth]{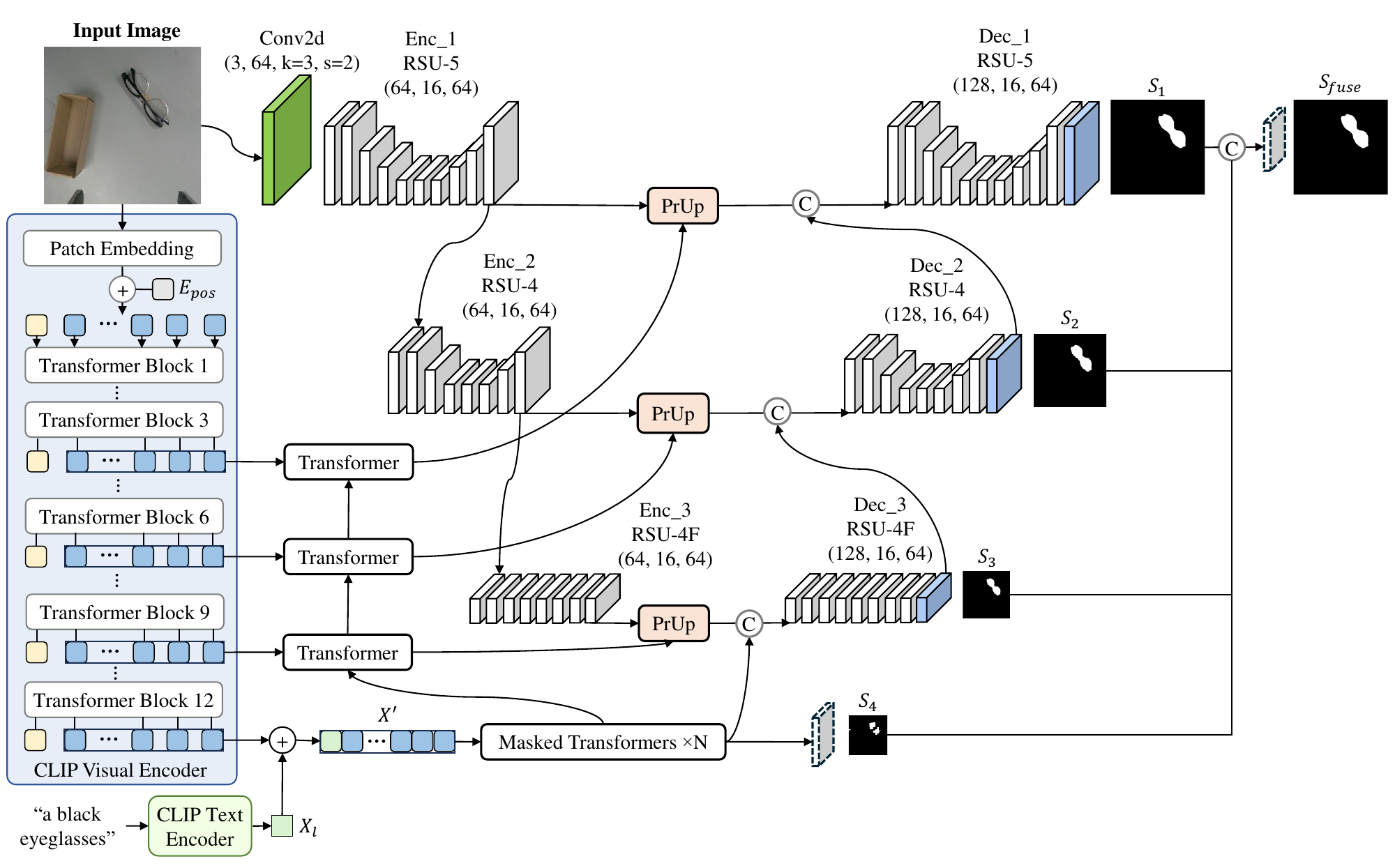}
\caption{The architecture of CLIPU$^2$Net. }
\label{fig:clipu2net}
\end{figure*}

\subsection{Network Architecture}
The architecture of CLIPU$^2$Net, visualized in Figure \ref{fig:clipu2net}, consists of three components: CLIP encoders to extract joint visual-text representations; a learnable masked multimodal fusion block that replaces Feature-wise Linear Modulation (FiLM); and a saliency module with U-squared decoder blocks for referring segmentation.

\textbf{CLIP Encoding}
Given an image and text, the CLIP text and vision encoders compute the embedded text and image token features as $X_l \in R^{D_l}$ and $[X^{[CLS]}_v; X_v] \in R^{(L+1) \times D_v}$, where $X^{[CLS]} \in R^{D_v}$ is the image [CLS] token.

\textbf{Masked Multimodal Fusion}
Previous studies \cite{cheng2022masked, lai2024eye} have observed that Transformer-based models often have slow convergence due to processing context across tokens globally before slowly attending to local regions. We hypothesize that this issue also affects multimodal fusion between text and image features using Transformer-based models. To address this, we propose to use masked attention. We linearly project $X_l$ and $X_v$ into a joint embedding space and fuse them into a multimodal feature, $X^{'} \in R^{(L+1) \times D}$:

\begin{equation} \label{clip_image}
\begin{split}
X^{'}_{l} = {Proj}_{l}(X_l), X^{'}_{v} = {Proj}_{v}(X_v) \\
X^{'} = [X^{'}_{l}; X^{'}_{v}] \\
\end{split}
\end{equation}

\noindent Replacing [CLS] token by the embedded token from text modality enforces multimodal encoder to pay more attention to the image tokens correlated to the context captured by text token. Furthermore, we composed of $N$ Transformer layers with masked attention, where the masked self-attention mechanism is computed as:

\begin{equation} \label{clip_image}
\begin{split}
X^{'}_{n+1} &= softmax((Q_{n} K_n^T + M_{n})/\sqrt{D})V_n \\
M_{n}(i,j) &= 
\begin{cases}
    0, & \text{if } i = j \text{ or } j = 1, \\
    -\infty, & \text{otherwise}. \\
\end{cases}
\end{split}
\end{equation}

\noindent where $n$ is the layer index, $Q_{n}$, $K_n$, $V_n$ $\in R^{(L+1)\times D}$ are the query, key, value features computed from the previous feature $X^{'}_n$. The attention mask $M_n$ ensures that the self-attention mechanism ignores interactions among image tokens, computing attention only between text tokens and image tokens and the self-attention of each token. The final multimodal feature $X^{'}_N$ is used to generate one side output $S_4$.

\textbf{Segmentation Decoder}
The segmentation decoder integrates a miniature U$^2$Net alongside the U-shaped transformer decoder to address the loss of multi-scale information caused by patch embedding in the CLIP visual encoder. The miniature U$^2$Net comprises 6 blocks, each consisting of stacks of Residual U-blocks (RSU). The U$^2$Net encoder processes the image input of size $H \times W \times 3$ and generates three multi-scale features ${F^1_e, F^2_e, F^3_e}$ with three RSU encoding blocks, where each feature $F^i_e \in \mathbb{R}^{H/2^i \times W/2^i \times D}$.

Moreover, we integrate the U-shape transformer encoder from the original CLIPSeg to allow early visual information to flow into U$^2$Net decoding process. Starting with the visual token feature $X^9_v \in \mathbb{R}^{L \times D_v}$ from the 9th CLIP visual encoding layer, it is processed by a transformer block, and reshaped into a 4D tensor $X'^9_v$ of $D$ dimension. Then, $X'^9_v$ and $F^3_e$ are taken by a Projection-Upsample (PrUp) block, where the low-resolution token feature is upsampled with bilinear interpolation and concatenated with the high-resolution feature $F^3_e$, before being processed by a RSU block, generating the new multi-scale feature $F^{'3}_e$. Next, $F^{'3}_e$ is concatenated with the final multimodal feature $X^{'}_L$ and processed by an RSU block, producing the decoding feature $F^{3}_d$ and the third side output $S_3$. This process is repeated for the visual token features from the sixth and third layers of the CLIP visual encoder, generating additional side outputs $S_2$, and $S_1$. All side outputs are concatenated and refined by a 1$\times$1 convolution, producing the final probability map $S_{fuse}$.

\textbf{Training Loss}
Similar to CLIPUNetr \cite{jiang2024clipunetr}, the total loss is calculated as the sum of the losses from all outputs:

\begin{equation} \label{loss_sum}
\begin{split}
\mathcal{L} = {\alpha}_{fuse} {\ell}_{fuse} + \sum_{k=1}^{4} ({\alpha}^k_{side} {\ell}^k_{side})
\end{split}
\end{equation}

\noindent where ${\ell}_{fuse}$ and ${\ell}^k_{side}$ denote the loss for the fused and the side probability maps. ${\alpha}_{fuse}$ and ${\alpha}^k_{side}$ are the weights for each loss component, empirically chosen as 1. Each loss term $\ell$ is computed as the sum of Focal and DICE loss.

\subsection{Geometric Constraints in Salient Vision}
Geometric constraints \cite{hespanha1999tasks} use points and lines to describe the alignment of a robot end effector to a target. Following \cite{gridseth2016vita}, four base geometric constraints are used:

\begin{equation} \label{visual_tasks}
\begin{aligned}
e_{pp}(\textbf{f}) &= f_2 - f_1 \\
e_{pl}(\textbf{f}) &= f_1 \cdot f_{34} \\
e_{ll}(\textbf{f}) &= f_1 \cdot f_{34} + f_2 \cdot f_{34} \\
e_{par}(\textbf{f}) &= f_{12} \times f_{34} \\
\end{aligned}
\end{equation}

\noindent where $e_{pp}$, $e_{pl}$, $e_{ll}$, and $e_{par}$ are denoted as the error signals for point-to-point (p2p), point-to-line (p2l), line-to-line (l2l), and parallel-line (par) constraint, respectively. A line $f_{ij}$ is computed as the cross product of two points, $f_i$ and $f_j$. Given the error signals $\dot{e}$ for the current visual observation, the UIBVS controller is governed by the visuomotor control law:

\begin{equation} \label{vs_error}
\dot{e} = J_{u}(q)\dot{q}
\end{equation}

\noindent where $\dot{q}$ is the control input of a robot with $N$ degrees-of-freedom, $J_{u}$ is the visuomotor Jacobian that maps the visual error signals to robot motions, and is updated online with Broyden's method \cite{jagersand1997experimental}. From the definitions, two problems need to be addressed: 

\begin{itemize}
\item What constraints should be used for the current task?

\item Which points and lines should be used to dynamically form the error signals of the constraints?

\end{itemize}

\noindent To tackle these problems, we first establish a baseline method to automatically determine the appropriate geometric constraints by the initial task observations. Next, we define attention interactions to select the appropriate pairings of the points and lines in real time.

\textbf{Determining Geometric Constraints}
Inspired by \cite{liu2024moka} that LLMs can generate robust point-based affordance representations to initiate control, we further investigate if LLMs are capable of inferring both point and line-based affordance representations encapsulated by the geometric constraints. Besides manually defining constraints by users through an HRI interface \cite{gridseth2016vita, jiang2024clipunetr}, we now explore a baseline strategy to automatically determine these constraints using GPT-4o. Given only the initial visual observation of the workspace and a text describing the manipulation task goal, we prompt GPT-4o to infer the choice of geometric constraints $E = {e_i | e_i \in (e_{pp}, e_{pl}, e_{ll}, e_{par})}$, as well as the target prompt $l$ to infer CLIP$^2$Net.

\textbf{Attention Interaction}
For the eye-in-hand camera configuration, attention interactions can be broadly categorized into two types: 1) object-gripper interaction, and 2) object-object interaction. For object-gripper interaction, the goal is to manipulate the target using the end effector. Affordances in this interaction arise from single-span attentions focused on the pixel locations of the target. Following the approach in \cite{jiang2024clipunetr}, PCA is used to infer over the output saliency map $S_{fuse}$, extracting the target point $f_2$ for the point-to-point constraint, or line $f_{34}$ for point-to-line, line-to-line, or parallel-line constraints, respectively. A heuristic static point (e.g. $f_1=(W/2, 4H/5, 1)$) or a vertical line $f_{12}$ passing through the mid-image center is used to complete the pairings.

For object-object interaction, where the object held by the end effector interacts with another target in the scene, affordances arise from dual attention spans: one over the object in the gripper and another over the target that completes the manipulation context. To generate geometric constraints, two prompts, $l_1$ and $l_2$, are provided to describe the carried object and the target. CLIPU$^2$Net is applied twice to generate two probability maps, $S^1_{fuse}$ and $S^2_{fuse}$, on which PCA is applied to derive pairs of points and lines. Since the carried object appears static from the eye-in-hand camera’s perspective, constraints are computed once for the carried object and dynamically updated for the target during control.






\section{Experiments}
\subsection{Experimental Settings}
\textbf{Referring Image Segmentation}
To evaluate the robustness of the proposed CLIPU$^2$Net in pixel-wise localization as well as boundary prediction quality, we use three datasets: 

\begin{itemize}
\item \textbf{PhraseCut} dataset \cite{wu2020phrasecut} contains 340,000 referring phrases with associating regional segmentation masks. We follow its evaluation protocol and report the Mean intersection-over-union (mIoU) and cumulative intersection-over-union (cIoU) metrics.

\item \textbf{UMD+GT} dataset \cite{xu2021affordance} contains 30,000 RGBD images of 104 objects in multiple views and 6 affordance labels. Following \cite{jiang2024clipunetr}, the labels are enriched with object-oriented and affordance-enriched prompts, and we report 4 metrics to measure prediction quality: Mean Absolute Error (MAE), structure measure ($S_\alpha$), weighted F-measure ($wF_{\beta}$), and max F-measure ($F^m_{\beta}$).

\item \textbf{DIS5K} dataset \cite{qin2022highly} contains 5,470 images of 225 categories. Each image is manually annotated with highly accurate segmentation mask outlining structural complexities of the objects. For experiments, we use the category of the object as the prompts, and report the 6 metrics measuring the quality of the predicted object boundary and structure.

\end{itemize}

\textbf{Robot Control}
We study the robot control system across four manipulation contexts: 1) Reach and grasp; 2) Pick and place; 3) Pull-open; and 4) Pour. 46 task targets with varying appearances are used. For each target, the robot is randomly positioned to view the target top-down or frontally. Two components are evaluated:

\begin{itemize}
\item \textbf{Correctness of Constraints:} For each manipulation target, a set of ground truth geometric constraints are selected by two skilled users. We assess the correctness of the baseline GPT-4o method to automatically determine the constraints, and report the accuracy. 

\item \textbf{Real-time Control:} Using the determined constraints, we enact real-world robot control with CLIPU$^2$Net-augmented perception vs. classical perception, where visual tracking is used following \cite{gridseth2016vita}. Three attempts are allowed to complete a task with one target. One failed attempt results in a success rate of 50\%, while two failed attempts result in a success rate of 0\%. Success rates are averaged across all targets. 

\end{itemize}

\textbf{Implementation Details}
CLIPU$^2$Net is implemented in PyTorch and trained on a single Nvidia Titan XP GPU with a batch size of 64, cosine annealing with an initial learning rate of 0.0005. The UIBVS control is implemented in ROS, utilizing Cartesian and joint spaces with support of velocity control for tabletop and front manipulation tasks. In comparison experiments, waypoint control is employed, moving the end effector incrementally until convergence.

\subsection{Results on Referring Image Segmentation}
\textbf{Quantitative Evaluation} 
Table \ref{table:phrasecut} and \ref{table:dis} present the quantitative results of referring image segmentation on the PhraseCut, UMD+GT, and DIS5K datasets, respectively. Our model achieves superior performance compared to MDETR \cite{kamath2021mdetr}, with only one-quarter of its size, demonstrating the benefits of using masked attention fusion for multimodal learning. Moreover, our model outperforms CLIPUNetr \cite{jiang2024clipunetr} in predicting fine boundaries and structures with fewer parameters, highlighting the effectiveness of the decoder design.

\begin{table}[h]
\centering
\caption{Quantitative results on PhraseCut and UMD+GT datasets.}
\label{table:phrasecut}
\begin{adjustbox}{width=\columnwidth,center}
\begin{tabular}{lccccccc}
\toprule
 & \multicolumn{2}{c}{\textbf{PhraseCut}} & \multicolumn{4}{c}{\textbf{UMD+GT}} \\
\cmidrule(lr){2-3} \cmidrule(lr){4-7}
\textbf{Model} & mIoU & cIoU & MAE & $S_\alpha$ & $wF_{\beta}$ & $F^m_{\beta}$ \\

\midrule

HulaNet~\cite{wu2020phrasecut} & 0.413 & 0.508 & - & - & - & - \\
MDETR~\cite{kamath2021mdetr} & 0.531 & 0.546 & - & - & - & - \\
CLIPSeg~\cite{luddecke2022image} & 0.461 & 0.562 & - & - & - & - \\
CLIPSeg (PC+)~\cite{luddecke2022image} & 0.434 & 0.547 & 0.134 & 0.663 & 0.384 & 0.566 \\
CLIPUNetr~\cite{jiang2024clipunetr} & 0.498 & 0.579 & 0.003 & 0.897 & 0.777 & 0.802 \\
AffKp~\cite{xu2021affordance} & - & - & 0.004 & 0.876 & 0.675 & 0.803 \\

\midrule
\textbf{CLIPU$^2$Net} & 0.542 & 0.589 & - & - & - & - \\
\textbf{CLIPU$^2$Net-Mix} & \textbf{0.544} & \textbf{0.591} & \textbf{0.001} & \textbf{0.932} & \textbf{0.843} & \textbf{0.866} \\
\bottomrule
\end{tabular}
\end{adjustbox}
\end{table}

\begin{table}[h]
\centering
\caption{Quantitative results on DIS5K test set.}
\label{table:dis}
\begin{adjustbox}{width=\columnwidth,center}
\begin{tabular}{lccccccc}
\toprule
 & \multicolumn{6}{c}{\textbf{DIS-TE (1-4)}} \\
\cmidrule(lr){2-7}
\textbf{Model} & $F^m_{\beta}$ & $F^w_{\beta}$ & MAE & $S_\alpha$ & $E_\phi^m$ & $HCE_\gamma$ \\

\midrule

BASNet~\cite{qin2019basnet} & 0.752 & 0.663 & 0.086 & 0.783 & 0.835 & 1313 \\
U$^2$Net~\cite{qin2020u2} & 0.761 & 0.670 & 0.083 & 0.791 & 0.835 & 1333 \\
ISNet~\cite{qin2022highly} & 0.799 & 0.726 & 0.070 & 0.819 & 0.858 & 1016 \\
FP-DIS~\cite{zhou2023dichotomous} & 0.831 & 0.770 & 0.057 & 0.847 & 0.895 & - \\
UDUN~\cite{pei2023unite} & 0.831 & 0.772 & 0.057 & 0.844 & 0.892 & \textbf{977} \\

\midrule
\textbf{CLIPU$^2$Net} & \textbf{0.846} & \textbf{0.779} & \textbf{0.054} & \textbf{0.858} & \textbf{0.899} & 1326 \\
\bottomrule
\end{tabular}
\end{adjustbox}
\end{table}

\begin{figure}[h!]
	\centering 
	\includegraphics[width=\columnwidth]{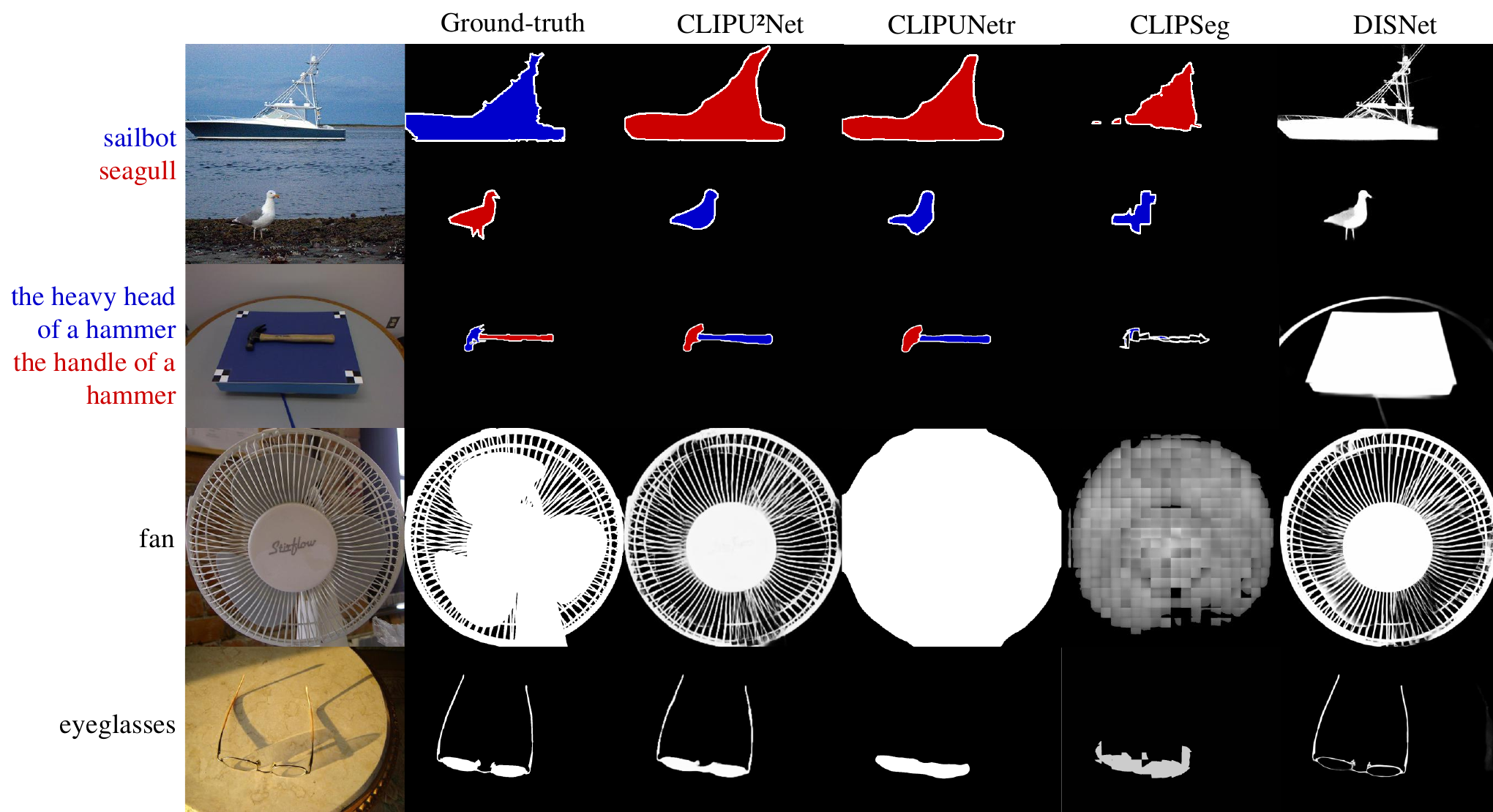}	
	\caption{Qualitative results for referring image segmentation.} 
	\label{fig:qualitative}%
\end{figure}

\textbf{Qualitative Evaluation}
Figure \ref{fig:qualitative} shows the prediction results of CLIPU$^2$Net compared to other methods. A common challenge in salient object segmentation is the uncertainty of what constitutes saliency, which should be adaptable based on a user's attention. Consequently, saliency models like DISNet \cite{qin2022highly} can struggle with localizing the full targets, while our model benefits from integrating CLIP for joint image-text representations to determine saliency and thus, achieving better localization capabilities. Our model also demonstrates comparable performance vs. other methods like CLIPSeg \cite{luddecke2022image}, avoiding checkerboard artifacts and having finer structures.

\textbf{Ablation Study} We validate the effectiveness of each key component in CLIPU$^2$Net by progressively removing the masked attention, U$^2$Net decoder, or both. Additionally, we construct a simplified baseline model, referred to as CLIPU$^2$Net-Bare. Similar to ViTSeg \cite{luddecke2022image}, the baseline model includes only a single convolution layer after the masked attention. The results are shown in Table \ref{Table_ablation}. Overall, both the masked attention and U$^2$Net decoder significantly contribute to learning fine-grain segmentation. The CLIPU$^2$Net-Bare model also demonstrates considerably higher performance compared to ViTSeg, highlighting the effectiveness of the masked attention.

\begin{table}[h!]
\caption{Ablation study with PhraseCut dataset.}
\label{Table_ablation}
\centering
\begin{adjustbox}{width=\columnwidth,center}
\begin{tabular}{l c c c c} 
 \hline

Model & No Masked Attn & Coarse & Extreme & mIoU \\ 
 \hline
CLIPU$^2$Net-v1 & \checkmark & - & - & 0.540 \\
CLIPU$^2$Net-v2 & - & \checkmark & - & 0.520 \\
CLIPU$^2$Net-v3 & \checkmark & \checkmark & - & 0.515 \\
CLIPU$^2$Net-Bare & - & \checkmark & \checkmark & 0.449 \\
ViTSeg \cite{luddecke2022image} & \checkmark & \checkmark & \checkmark & 0.389 \\

\hline
\end{tabular}
\end{adjustbox}
\end{table}

\subsection{Results on Robot Control}

\textbf{Quantitative Evaluation}
We report the average success rate and comparison against classical Vita control \cite{gridseth2016vita} in Table \ref{Table:result_robot}, as well as the accuracy of the predicted geometric constraints. With the incorporation of CLIPU$^2$Net, our robot control system achieves consistent performance in performing manipulation tasks with various targets. On the other hand, for contexts like pick-and-place, the classical interface struggles significantly. This is due to the fact that parts of the objects are obscured as the eye-in-hand camera moves, making the annotations and the tracking of the targets substantially harder. Moreover, the classical interface struggles to track transparent or more reflective objects like glass cups. 

Additionally, GPT-4o shows some innate abilities in predicting line-based constraints from visual observations, especially for tasks involving part affordances like handles. This further validates the fact that manipulation contexts can arise from image geometry, which can be captured using point and line-based affordance representations. Still, the prediction quality varies across tasks, indicating some challenges in handling diverse contexts. 



\begin{table*}[t]
\centering
\caption{Results for robot control.}
\label{Table:result_robot}
\begin{adjustbox}{center}
\begin{tabularx}{\textwidth}{lXcccccc}
\toprule
\multicolumn{4}{c}{} & \multicolumn{2}{c}{\textbf{Success Rate}} \\
 \cmidrule(lr){5-6}
\textbf{Context} & \textbf{Target} & \textbf{Constraint(s)} & \textbf{Acc} & CLIPU$^2$NETR & Classical \cite{gridseth2016vita} \\
\midrule
Reach-and-Grasp & Apple. lemon, red pepper, carrot, banana, umbrella, tennis ball, 4 marker pen. & p2p, p2l, l2l, par & 100\% & 100\% & 91\% \\

 & 4 beverage can, 2 beverage bottles, 2 water bottles, med-bottle, 2 spray bottle. & p2l, p2p & 72.7\% & 100\% & 91\% \\

 & Plastic cup, glass cup. & par, p2l, p2p & 50\% & 75\% & 38\% \\

 & Grasp-Fork-in-Plate. Grasp-Spoon-handle-in-Bowl. & par, p2p & 100\% & 100\% & 50\% \\

 & The handle of a screwdriver, 2 hammers, 2 knife. & par, p2p & 90\% & 60\% & 100\% \\

\midrule

Pick-and-Place & Strawberry, lemon $\rightarrow$ Bowl; Apple, carrot  $\rightarrow$ Plate. & par, p2p; p2p & 87.5\% & 100\% & 75\% \\

 & Coke can $\rightarrow$ Basket; Beverage can $\rightarrow$ Bucket. & p2p; p2p & 100\% & 100\% & 75\% \\

 & Tea bag $\rightarrow$ Tea cup, 2 Mugs. & p2p; p2p & 91.7\% & 83\% & 83\% \\

 & Eyeglasses $\rightarrow$ Eyeglass box; Marker pen $\rightarrow$ Rectangle basket. & par, p2p; par, p2p & 50\% & 100\% & 67.5\% \\

\midrule

Pull-open & Left and right closet handles. & l2l, p2p & 50\% & 100\% & 0\% \\
 & Drawer handle. & p2l, p2p & 50\% & 100\% & 100\% \\

\midrule

Grasp-and-Pour & Pepper-in-bottle $\rightarrow$ Plate & p2l, p2p; p2p & 100\% & 100\% & 0\% \\

\bottomrule
\end{tabularx}
\end{adjustbox}
\end{table*}


\begin{figure*}[ht!]
\centering
\includegraphics[width=2\columnwidth]{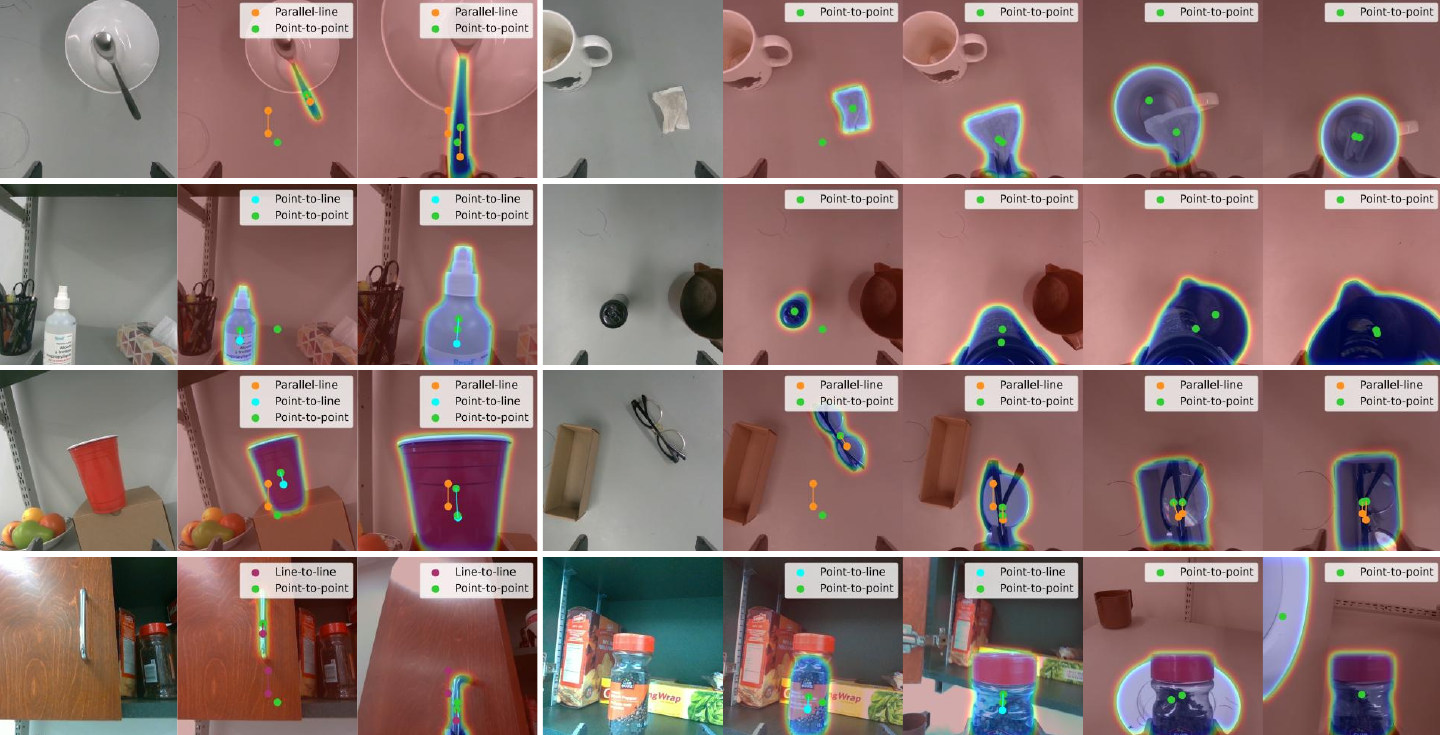}
\caption{Results of the predicted geometric constraints and motions for 8 of the 46 assessed tasks. }
\label{fig:result_robot}
\end{figure*}

\textbf{Qualitative Evaluation}
Figure \ref{fig:result_robot} visualizes the geometric constraints for 8 out of the 46 assessed tasks. With the integration of CLIPU$^2$Net, we can segment affordance regions, successfully completing tasks requiring fine-grain affordance information like reaching for the spoon handle in a bowl and placing a teabag into a mug, as seen in the first row. CLIPU$^2$Net also handles diverse contexts effectively, from segmenting door handles for pulling tasks in the fourth row, to segmenting cans from top-down or frontal views for pick-and-place tasks in the third row. Additionally, the usage of line-based constraints ensures stable motions, such as in the eyeglasses placement task (third row, fourth column), where the parallel-line constraint allow safe placement of the glasses into the tiny space of a box. In summary, geometric constraints are universal and effective visual representations of motion, adaptable to various contexts.

\begin{figure}[h!]
	\centering 
	\includegraphics[width=\columnwidth]{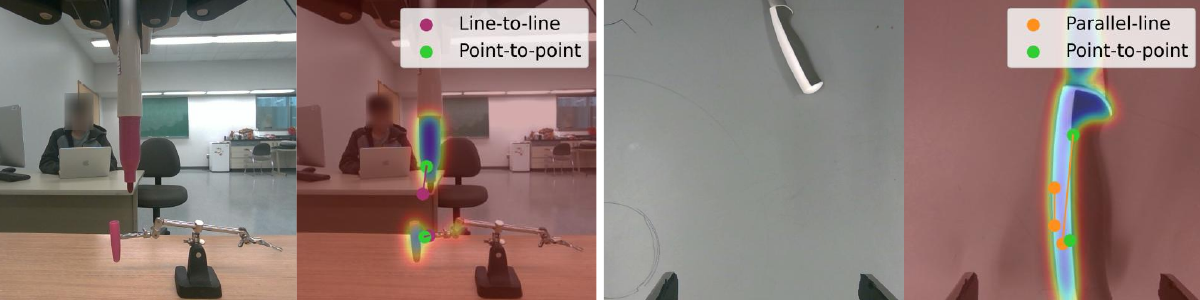}	
	\caption{Some failure cases.} 
	\label{fig:FAILURE}%
\end{figure}

\textbf{Limitations} An issue with CLIPU$^2$Net is its limitations in handling part segmentation, which affects its performance in tasks requiring detailed differentiation of object parts. We provide two failed tasks resulted from this limitation in Figure \ref{fig:FAILURE}. 
In the first task, the model fails to segment the cap of the marker pen when prompted with "pen cap". In the second task, the model fails to distinguish the blade from the handle of the knife. We hypothesize that this limitation arises from the annotations in the PhraseCut dataset, which lacks part annotations. We plan to evaluate this further in future work.

\section{Conclusions}
In this paper, we introduce CLIPU$^2$Net, a compact referring image segmentation model integrated into a robot's perception module, and use it to extract salient visual features as geometric constraints. Experimental results validate the effectiveness of our approach in enacting real-world robot control with eye-in-hand visual servoing. For future work, we aim to explore part segmentation from referring expressions to further refine contextual understanding and improve manipulation accuracy. Furthermore, we plan to investigate how geometric constraints can be leveraged to enhance the learning of referring image segmentation models themselves, optimizing their performance in real-world robot manipulation environments.






{
\bibliographystyle{IEEEtranS}
\bibliography{citation}
}

\end{document}